\documentclass[runningheads]{llncs}
\usepackage{graphicx}
\usepackage{array}
\usepackage{algorithm2e} 
\usepackage{subcaption}
\usepackage{floatrow}
\newfloatcommand{capbtabbox}{table}[][\FBwidth]

\usepackage{listings}
\usepackage{xcolor}
\usepackage{multirow}

\colorlet{punct}{red!60!black}
\definecolor{background}{HTML}{EEEEEE}
\definecolor{delim}{RGB}{20,105,176}
\colorlet{numb}{magenta!60!black}

\lstdefinelanguage{json}{
    basicstyle=\normalfont\ttfamily,
    numbers=left,
    numberstyle=\scriptsize,
    stepnumber=1,
    numbersep=8pt,
    showstringspaces=false,
    breaklines=true,
    frame=lines,
    backgroundcolor=\color{background},
    literate=
     *{0}{{{\color{numb}0}}}{1}
      {1}{{{\color{numb}1}}}{1}
      {2}{{{\color{numb}2}}}{1}
      {3}{{{\color{numb}3}}}{1}
      {4}{{{\color{numb}4}}}{1}
      {5}{{{\color{numb}5}}}{1}
      {6}{{{\color{numb}6}}}{1}
      {7}{{{\color{numb}7}}}{1}
      {8}{{{\color{numb}8}}}{1}
      {9}{{{\color{numb}9}}}{1}
      {:}{{{\color{punct}{:}}}}{1}
      {,}{{{\color{punct}{,}}}}{1}
      {\{}{{{\color{delim}{\{}}}}{1}
      {\}}{{{\color{delim}{\}}}}}{1}
      {[}{{{\color{delim}{[}}}}{1}
      {]}{{{\color{delim}{]}}}}{1},
}

\begin{document}
%
\pagenumbering{gobble}
\pagestyle{empty}

\title{Enhancing Legal Document Retrieval:\\ A Multi-Phase Approach with \\ Large Language Models}
\titlerunning{Multi-Phase LLM-based Legal Document Retrieval}
%
\author{Hai-Long Nguyen $^*$\inst{1} \and
Duc-Minh Nguyen $^*$ \inst{2} \and
Tan-Minh Nguyen\inst{1} \and \\
Ha-Thanh Nguyen\inst{3} \and
Thi-Hai-Yen Vuong\inst{1} \and
Ken Satoh\inst{3}
}
\authorrunning{Nguyen et al.}
%
\institute{VNU University of Engineering and Technology, Hanoi, Vietnam \\
\email{long.nh@vnu.edu.vn}  \and
RMIT University, Vietnam\and
National Institute of Informatics, Tokyo, Japan}
 \maketitle              

\def\thefootnote{*}\footnotetext{These authors contributed equally to this work}

\begin{abstract}
Large language models with billions of parameters, such as GPT-3.5, GPT-4, and LLaMA, are increasingly prevalent. Numerous studies have explored effective prompting techniques to harness the power of these LLMs for various research problems. Retrieval, specifically in the legal data domain, poses a challenging task for the direct application of Prompting techniques due to the large number and substantial length of legal articles. This research focuses on maximizing the potential of prompting by placing it as the final phase of the retrieval system, preceded by the support of two phases: BM25 Pre--ranking and BERT--based Re--ranking. Experiments on the COLIEE 2023 dataset demonstrate that integrating prompting techniques on LLMs into the retrieval system significantly improves retrieval accuracy. However, error analysis reveals several existing issues in the retrieval system that still need resolution.
\end{abstract}
\section{Introduction}

In a rapidly developing digital world, there are enormous amounts of information and queries that need to be processed every minute. Therefore, there is a demand for an effective and efficient search engine that would free law practitioners from heavy manual work. Legal information retrieval is a challenge that has gained interest from both researchers and industry recently. This task study is about the automation of retrieving accurate pieces of information for a given query from users. 

In this study, we focus on Legal Article Retrieval, an essential task for many countries that follow a statute law system. The primary objective is to develop a retrieval system that can effectively pinpoint a subset of legal articles that pertain to a given user query. Let $Q$ denote the initial query and $C$ represents a corpus of legal articles (i.e., $C=\{a_1,a_2,\dots,a_n\}$). The main goal is to extract a subset $A \subseteq C$, where each $a_i \in A$ is relevant to the query $Q$. 
To achieve this, we propose an instructional  \textbf{relevance computation} approach, instructing Large Language Models (LLMs) to generate a set of relevance scores $S$ of a group of legal articles $A$ to a given query $Q$:
\begin{equation}
    S = LLMs(Q,A)
\end{equation}

The proposed prompting technique enables flexibility during 
ensembling with other retrieval models by offering relevance scores. 




This paper thoroughly discusses the implemented experiments and provides an in-depth analysis of the effectiveness, limitations, and real-world applications of our proposed approach. By examining both successful and unsuccessful cases, we aim to identify the strengths and weaknesses of the current methodology. Furthermore, we outline potential directions for future improvements, modifications, and enhancements to the process of legal information retrieval, ultimately contributing to the development of a valuable tool for legal practitioners.




 
\section{Related Work}

Legal information retrieval is an important problem in juris-informatics. The methodologies have evolved considerably over time, with methods ranging from simple to complex, and from older to newer approaches \cite{rosa2021yes,yoshioka2022hukb,nguyen2023neco,nguyen2024captain}. Early studies focused on extracting relevant legal information using information retrieval (IR)-based approaches. For instance, Gao et al. (2019) employed topic keywords to retrieve legal documents relevant to a given situation by using an information retrieval model \cite{gao2019fire2019}. However, this method relies heavily on the extracted keywords, which may not be sufficient for representing complex legal texts.

As the research shifted toward using deep learning architectures, attention-based models were proposed to achieve better representations of legal texts. Nguyen et al. (2022) employed attentive deep neural networks to construct two hierarchical network architectures, Attentive CNN and Paraformer, to represent long sentences and articles in legal texts \cite{nguyen2022attentive}. These architectures utilized attention mechanisms to capture important information in lengthy legal documents.

Building upon these previous works, Hai-Long et al. (2023) introduced a joint learning model to improve the performance of information retrieval and entailment tasks on legal texts \cite{hai-long-etal-2023-joint}. This approach exploited the relationship between relevance and conclusion to obtain better results with multi--task learning.

Moreover, chain-of-thought prompting in large language models has been investigated, as demonstrated by Wei et al. (2022) \cite{wei2022chain}. This research leveraged the power of large attentive language models to display connected reasoning results, providing an improved way of eliciting reasoning in LLMs for legal text processing.

As the focus shifted to large language models, Sun et al. (2023) explored the potential of using generative LLMs like ChatGPT and GPT-4 for relevance ranking in IR \cite{sun2023chatgpt}. They showed that these large language models could achieve better performance than conventional supervised methods, even outperforming state-of-the-art models like monoT5-3B on various benchmarks.

\section{Prompting-supported Retrieval Pipeline}

\begin{figure}
    \centering
    \includegraphics[width=1\linewidth]{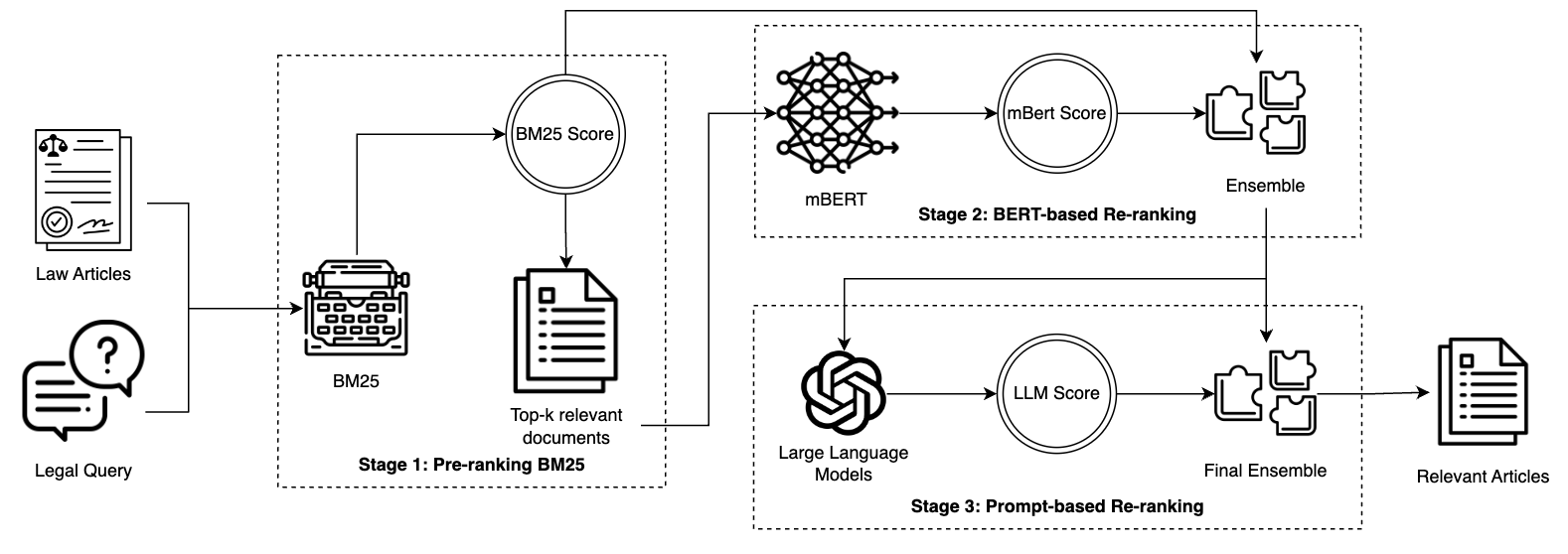}
    \caption{Prompting-supported Retrieval Pipeline.}
    \label{fig:pipeline}
\end{figure}
Due to the considerable number of articles present in the Legal Statute Corpus, ensuring efficient retrieval time and optimizing computational complexity necessitates the implementation of a multi-phase retrieval system to gradually filter candidates from the corpus. Drawing inspiration from the Retrieval pipeline proposed in \cite{hai-long-etal-2023-joint}, the search pipeline in this paper comprises three main phases: Pre--ranking utilizing BM25, Re--ranking phase employing a model based on the BERT architecture, and finally, another Re--ranking phase employing Prompting techniques on large language models (LLMs). 
Figure \ref{fig:pipeline} visualizes specific steps of the retrieval pipeline.
    
\subsection{Pre--ranking BM25}
\label{subsec:bm25-preranking}

The selected dataset for experimentation is derived from the COLIEE competition, encompassing a corpus of 768 articles\footnote{\url{https://www.japaneselawtranslation.go.jp/en/laws/view/3494/en}}. Given the substantial number of articles, the primary objective in the initial phase of the model is to ensure both speed and an adequate recall score to cover a significant portion of ground-truth articles. Further refinement to identify truly relevant documents will be conducted in subsequent phases. To meet these requirements, the Okapi BM25 ranking model \cite{robertson1995okapi} based on the lexical features is employed to evaluate the relevance score. Leveraging statistical analysis and token frequency computations, the BM25 model exhibits rapid execution and maintains the recall score irrespective of document length or query structure, making it suitable for the specified objectives.

\subsection{BERT-based Re--ranking}
\label{subsec:bertbased-reranking}

The utilization of the BM25 \cite{robertson1995okapi} model in the pre--ranking phase primarily relies on the lexical feature, emphasizing statistical token frequency without considering the semantic similarity between the query and legal articles. Semantic similarity, representing the meaning when combining tokens and putting them in a context, is a crucial feature in the retrieval system. It aids in eliminating documents that, despite lexical relevance to the query, lack semantic relevance. Incorporating semantic similarity can significantly enhance precision scores for the entire retrieval system.

\paragraph{Training the BERT-based Re--ranking model} To address semantic similarity, the model based on the BERT architecture with a multi--task learning design, as proposed in \cite{hai-long-etal-2023-joint}, has been employed in the re-ranking phase.
The arguments and experiments conducted in the aforementioned research have demonstrated that information regarding the correctness of queries and the relevance of a \textit{(query, legal document)} pair mutually support each other. Employing a multi-task model to predict both pieces of information enhances the efficiency of the training process and improves the model's ability to compute semantic correlations \cite{hai-long-etal-2023-joint}. Specifically, the input to the multi--task model consists of a sequence generated by concatenating the query and the article, separated by a special token [SEP]. The output will consist of two main components: a probability (ranging from $0$ to $1$) indicating the level of semantic relevance between the query and the article, and another probability (ranging from $0$ to $1$) indicating whether the query is true or false based on the content of the article. The model will be trained based on two labels provided in the COLIEE dataset: relevant articles of the query and labels for the correctness of the query.

Given that the Legal Corpus data from the COLIEE workshop is derived from the Japanese Civil Code, the original language of the dataset is Japanese. Previous research \cite{hai-long-etal-2023-joint,vuong2024nowj,nguyen2024captain} has consistently demonstrated higher performance when experimenting with Japanese datasets compared to English. Therefore, the pre--trained parameters of Multilingual--BERT\footnote{\url{https://huggingface.co/bert-base-multilingual-cased}}, trained on a multilingual dataset including Japanese, are utilized. The utilization of a multilingual language model also facilitates future research and comparisons when employing pipelines with other legal datasets containing various languages.

\paragraph{Ensemble relevance score and inference} After the training process, the model will then be utilized for inference in the re-ranking phase. To leverage the strengths of both the BM25 and Multi-Task BERT models, the relevance scores from the two models will be ensemble to calculate the correlation score for the current re-ranking phase. The equation \ref{eq:ensemble-bert-bm25} describes how to combine the relevance scores of the BM25 model and the BERT model. 


\begin{equation}
    \label{eq:ensemble-bert-bm25}
    R_{reranking-phase-1} = \alpha * BM25\_score + \beta * BERT\_score
\end{equation}
whereas $\alpha$ and $\beta$ are weighting factors for the two models, $BM25\_score$ represents the correlation score calculated by the BM25 model, and $BERT\_score$ represents the correlation score calculated by the BERT model. The final correlation score $R_{reranking-phase-1}$ will be normalized to a range from $0$ to $1$. Afterwards, all query-article pairs with a correlation score $R_{reranking-phase-1}$ greater than a predefined $threshold1$ will be considered relevant and added to the candidate set for the next phase. An optimal hyperparameter set of ($\alpha$, $\beta$, $threshold1$) will be determined through a grid-search process on the validation set. 

The objective of the re-ranking phase using the BERT model is to filter articles that have semantic relevance to the query while still maintaining a relatively high number of relevant candidates according to the ground truth. Specifically, the goal was to retain an F2 score greater than or equal to $0.5$ while achieving the highest possible recall score. 

\subsection{Prompting-based Re-ranking}
\label{subsec:llm-bert-reranking}
In the previous retrieval phases, models mainly relied on lexical and semantic similarity. However, in reality, queries may not always directly address the core issue and may require logical reasoning to find relevant legal articles related to the query. Large language models (LLMs) are deep learning models with billions of parameters, trained on vast amounts of data. One notable point is that LLMs can learn to handle downstream tasks through instruction inputs.
Therefore, applying Prompting techniques to LLMs can leverage their basic-level reasoning capabilities. Additionally, their broad domain knowledge can help LLMs perform well with queries that do not use specialized legal terminology.

Due to the relatively long length of legal texts, and considering the token limit for input in large language models, typically around 25000 words, we have opted to explore the Zero-shot prompting method to preserve the complete content of both the query and the candidates. The Few-shot method, wherein the LLMs are provided with a set of samples for learning, will be discussed and investigated in detail in subsequent studies.

The input content for the prompting method proposed will include the query content, as well as the content and identifiers (e.g. Article 705, Article 01, ...) of all candidate articles obtained from the previous re-ranking phase using the Multi-Task BERT model. The output of the prompting sentences will include relevance scores for candidates ranging from $0$ to $100$. The output format specified in the prompting sentence will adhere to JSON format.


\paragraph{Adaptive Sliding window}
\begin{figure}
    \centering
    \includegraphics[width=1\textwidth]{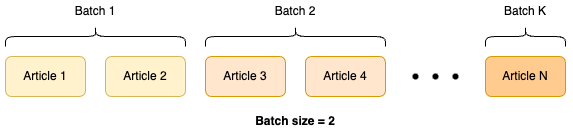}
    \caption{Visualization of sliding window technique.}
    \label{fig:adaptive-sliding-window}
\end{figure}
To encompass the complete content of as many candidates as possible within a single prompt, the adaptive sliding window prompting technique has been employed. Specifically, the content of candidates is included in the prompt in its entirety until approaching the token limit, at which point the process is halted. Employing this adaptive sliding window prompting technique minimizes the need for subdividing the prompt for a single query, allowing the LLMs to consider as many candidates simultaneously as possible. An example of this technique is illustrated in Figure \ref{fig:adaptive-sliding-window}.

\paragraph{Ensemble with BERT re-ranking relevance score.} 

Since each LLMs model as well as BERT-based re-ranking has its own strengths and weaknesses, the technique of ensembling the relevance scores of each model is carried out and evaluated for effectiveness based on the F2 score of the entire pipeline.To leverage the strengths of both models, their relevance scores are combined with weights using the formula \ref{eq:ensemble-bert-llm}.

\begin{equation}
    \label{eq:ensemble-bert-llm}
    R_{reranking-phase-2} = \beta* BERT\_score + \gamma * LLM\_score
\end{equation}
whereas $BERT\_score$ represents the correlation score of the BERT model, and $LLM\_score$ represents the correlation score obtained from the prompting results. $\beta$ and $\gamma$ are the respective weights assigned to the BERT and LLMs. Similar to the previous phase, $threshold2$ is used to determine which documents are ultimately considered relevant. Likewise, a grid-search process will be conducted on the validation set to identify the optimal set of hyper--parameters.


\section{Experiment and Results}
\subsection{Datasets}
COLIEE is an annual competition with the main purpose is nurturing a worldwide legal research community. The competition involves different challenges, covering both case law and statute law systems. The statute law database uses the 768 articles in the Japanese Civil Code. Meanwhile, original queries are questions selected from the Japanese Bar Exam and come along with an English translation version. Every year, COLIEE organizers select new questions from the Exam Bar to construct a new testing set, while testing and training sets of the previous year are merged to form a new training set. 

\begin{figure}[ht]
\centering
  \begin{subfigure}[b]{0.5\textwidth}
    \includegraphics[width=\textwidth]{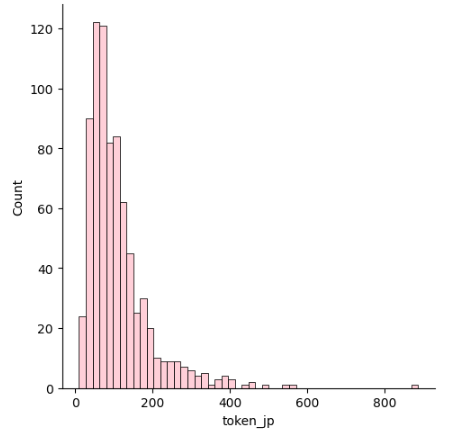}
    \caption{Legal article corpus.}
    \label{fig:f1}
  \end{subfigure}
  \begin{subfigure}[b]{0.45\textwidth}
    \includegraphics[width=\textwidth]{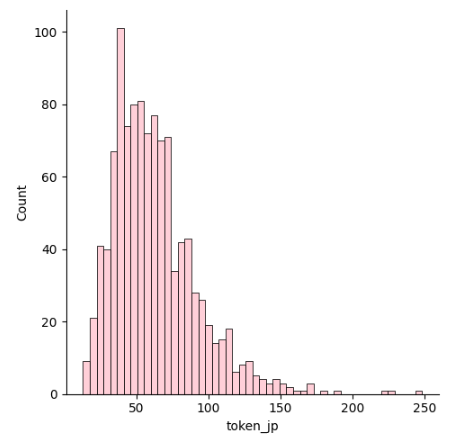}
    \caption{Legal query corpus}
    \label{fig:f2}
  \end{subfigure}
  \caption{Length distribution statistics of the COLIEE 2023.}
  \label{fig:query-histogram}
\end{figure}

In this work, we validate the proposed pipeline on the COLIEE 2023 dataset. Table \ref{tab:statistics-article} presents the statistics of articles' length in the dataset. On average, each article contains 109 tokens in Japanese and 100 tokens in English. Articles' length focuses on the range of 75 to 125 tokens, and can vary significantly from 10 tokens to more than 800 tokens as shown in Figure \ref{fig:f1}. For statistics of the query corpus, details are reported in Table \ref{tab:statistics-query}. There are 996 samples in the training set and 101 queries in the testing set, with the range of length starting from 13 tokens to 248 tokens. Figure \ref{fig:f2} illustrates the length distribution of queries in the dataset. Most queries contain around 30 to 70 tokens, though some queries have extreme lengths with more than 200 tokens. 

\begin{table}[ht]
    \caption{Statistics of articles' length in COLIEE 2023.}
    \centering
    \begin{tabular}{l|ccc}
    \hline
             & \textbf{Min} & \textbf{Max} & \textbf{Avg} \\ \hline
    Japanese & 10.0         & 886.0        & 109.6         \\
    English  & 5.0          & 867.0        & 100.2         \\ \hline
    \end{tabular}
    \label{tab:statistics-article}
\end{table}

    

\begin{table}[ht]
\centering
\caption{Statistics of queries' length in COLIEE 2023.}
\label{tab:statistics-query}
\begin{tabular}{|l|c|c|}
\hline
                          & \textbf{Training set} & \textbf{Testing set} \\ \hline
Number of queries         & 996                   & 101                  \\ \hline
Minimum length of a query & 13.00                 & 25.00                \\ \hline
Maximum length of a query & 248.00                & 130.00               \\ \hline
Average length of a query & 62.21                 & 65.28                \\ \hline
\end{tabular}
\end{table}

\subsection{Experiment Setup}

To achieve optimal efficiency, models in each retrieval phase need to be fine-tuned with appropriate parameters to accomplish the goals of each retrieval phase best. During the fine-tuning and pipeline construction process, several hyperparameters need to be adjusted to achieve the most effective search efficiency. Therefore, 20\% of the training set samples have been separated to form the validation set. After the training process, the hyperparameters will be adjusted based on the F2 score of the entire pipeline on the validation set. The specific deployment of models in each phase will be described in the following sections.

\paragraph{Pre--ranking with BM25}

As mentioned in Section \ref{subsec:bertbased-reranking}, the goal of the pre-ranking phase is to select documents with lexical relevance to the query. The expectation for the re-ranking phase is to maximize the removal of irrelevant legal texts while maintaining a relatively high recall score. After obtaining relevance scores from the BM25 model, the top-k documents with the highest relevance scores are chosen as candidates for the subsequent phases.

To select the optimal top-k, a recall score analysis of the BM25 model is conducted on the entire training dataset. Table \ref{tab:recall-bm25-score} presents the recall scores of the BM25 model for each top-k value.
\begin{table}[ht]
\centering
\setlength{\arrayrulewidth}{0.1mm}
\setlength{\tabcolsep}{0.01\textwidth}
\renewcommand{\arraystretch}{1.5}
\caption{Recall scores of the BM25 model.}
\begin{tabular}{c|c|c|c|c|c|c|c|c|c}
\textbf{Top-k}  & \textbf{500}    & 400    & 300    & 200    & 100    & 50     & \textbf{30}     & 10     & 5      \\ \hline
\textbf{Recall} & \textbf{0.9467} & 0.9317 & 0.9186 & 0.8931 & 0.8515 & 0.8050 & \textbf{0.7784} & 0.7076 & 0.6444
\end{tabular}
\label{tab:recall-bm25-score}
\end{table}
\noindent Based on the table \ref{tab:recall-bm25-score}, it can be observed that choosing a top-k of $500$ can eliminate approximately $34.90\%$ of legal documents from the total corpus while still maintaining a suitable recall score of $0.9467$.

\paragraph{Re--ranking with Multi-Task BERT-based model}

Based on the analysis from paper  \cite{hai-long-etal-2023-joint}, the training process of the Multi-Task BERT model involves using the top 30 candidates from BM25 to form the training set and utilizing the pre-trained parameters of Multilingual-BERT\footnote{\url{https://huggingface.co/bert-base-multilingual-cased}}. 

The model, after training, is then utilized for inference in the re-ranking phase. As mentioned in Section \ref{subsec:bertbased-reranking}, the optimal parameter set on the validation set ensures the specified requirements: ($\alpha = 0.17$, $\beta=0.83$, $threshold1=0.921$). The results obtained after the BERT-based re-ranking phase on the validation set are $F2=0.527$, $Precision=0.274$, and $Recall=0.921$. 
The first row of Table \ref{tab:experiment-result} describes the result of the retrieval pipeline that contains two phases BM25 Preranking and BERT-based Re-ranking.

\paragraph{Re--ranking with Prompting LLM}
The prompting process to obtain relevance scores will be sequentially conducted on three versions of large language models, namely: GPT-3.5-turbo-instruct, GPT-3.5-turbo-1106, and finally, GPT-4-1106-preview, utilizing the API provided by OpenAI\footnote{https://openai.com}. As described in section \ref{subsec:llm-bert-reranking}, a grid-search phase is executed on the validation set, resulting in the optimal set of hyperparameters being ($\beta=0.5$, $\gamma=0.5$, $threshold2=0.52$) for ensembling the relevance score of BERT-based reranking model and the score of prompting process.

\subsection{Experiment Result}

After the experimentation process, Table \ref{tab:experiment-result} describes the obtained results of the retrieval system with different settings and various large language models. The first row (1) of the table depicts the results of the retrieval system consisting of only two phases: pre--ranking with BM25 and re-ranking with the BERT-based model. Rows (2), (3), (4) and (5) show the results of the retrieval system with all three phases, where the final phase uses \textit{gpt-3.5-turbo-instruct} (for row 2), \textit{gpt-3.5-turbo-1106} (for row 3), and \textit{gpt-4.0-1105} (for both row 4 and 5). The pipeline's last phase used at row (2), (3) and (4) has the relevance scores only derived from LLMs. Meanwhile, the last phase of the pipeline used in row (5) has relevance scores that are ensembled from the  LLMs' relevance scores and re--ranking BERT's relevance scores (which were derived at previous phase).

\begin{table}[ht]
\centering
    \caption{Experiment results on COLIEE 2023.}
\begin{tabular}{|p{5cm}|ccc|ccc|}
\hline
\multicolumn{1}{|c|}{\multirow{2}{*}{\textbf{Retrieval System}}}                           & \multicolumn{3}{c|}{\textbf{Validation set}}                                                 & \multicolumn{3}{c|}{\textbf{Testing set}}                                                    \\ \cline{2-7} 
                                                           & \multicolumn{1}{c|}{\textbf{F2}} & \multicolumn{1}{c|}{\textbf{P}} & \textbf{R} & \multicolumn{1}{c|}{\textbf{F2}} & \multicolumn{1}{c|}{\textbf{P}} & \textbf{R} \\ \hline
BM25 + BERT-reranking (1)                                                   & \multicolumn{1}{c|}{0.5000}      & \multicolumn{1}{c|}{0.2620}             & \textbf{0.9116}          & \multicolumn{1}{c|}{0.5268}      & \multicolumn{1}{c|}{0.2738}             & \textbf{0.9207}          \\ \hline
gpt-3.5-turbo-instruct (2)  & \multicolumn{1}{c|}{0.5969}            & \multicolumn{1}{c|}{0.3600}                   &                 0.8674& \multicolumn{1}{c|}{0.6213}            & \multicolumn{1}{c|}{0.4084}                   &                 0.8514\\ \hline
gpt-3.5-turbo-1106 (3) & \multicolumn{1}{c|}{0.6649}            & \multicolumn{1}{c|}{0.5467}                   &                 0.7666& \multicolumn{1}{c|}{0.6620}            & \multicolumn{1}{c|}{0.4546}                   &                 0.8613\\ \hline
gpt-4-1106-preview (4)                    & \multicolumn{1}{c|}{0.7621}      & \multicolumn{1}{c|}{\textbf{0.7841}}             & 0.7799          & \multicolumn{1}{c|}{0.7647}      & \multicolumn{1}{c|}{\textbf{0.7564}}             & 0.7871          \\ \hline
gpt-4-1106-preview ensemble (5) & \multicolumn{1}{c|}{\textbf{0.7678}}      & \multicolumn{1}{c|}{0.7029}             & 0.8308          & \multicolumn{1}{c|}{\textbf{0.8085}}      & \multicolumn{1}{c|}{0.7277}             & 0.8712          \\ \hline
\end{tabular}

    \label{tab:experiment-result}
\end{table}

Comparing the retrieval system (1) with the remaining systems, it is evident that the prompting-based re-ranking phase significantly increases precision scores on both the test and validation sets. 
The retrieval systems (1), (2), and (3) make significant improvements of $0.1346$, $0.1808$, and $0.4826$ in terms of precision on the testing set compared to the baseline (1). 
It can be observed that the final retrieval phase does not considerably decrease the recall score, leading to a substantial improvement in the F2 scores of the query systems. 
Particularly, the F2 score of the retrieval system (4) is enhanced by $0.2379$ from $0.5268$ to $0.7647$ compared to the baseline (1). Indeed, this result indicates the effectiveness of our proposed retrieval pipeline in improving the accuracy of retrieved articles.

When comparing retrieval systems (4) and (5), it can be observed that combining the scores of the BERT-based re-ranking model with the scores from the LLMs prompting output, although slightly decreasing precision from $0.7564$ to $0.7277$, maintains recall scores from the two preceding phases. Specifically, the recall score of the retrieval system (5) is $0.8712$, which is higher by roughly 8\% compared to the retrieval system (4). Therefore, the F2 score of the retrieval system (5) improves by 4\% overall. 
Through this observation, it is evident that ensemble relevance scores of the two models have been effective in combining the strengths of both language models: the BERT-based model with the ability to understand semantic similarity between legal documents, and the LLMs with basic logical inference capabilities across a broad knowledge domain.



\begin{table}[ht]
\centering
\caption{Results of participating teams in the COLIEE 2023.}
\label{tab:compare-res}
\begin{tabular}{|lccc|}
\hline
\multicolumn{1}{|c|}{\textbf{Team}}               & \multicolumn{1}{c|}{\textbf{F2}} & \multicolumn{1}{c|}{\textbf{P}} & \textbf{R} \\ \hline
\multicolumn{1}{|l|}{CAPTAIN}                     & \multicolumn{1}{c|}{0.7645}            & \multicolumn{1}{c|}{0.7333}             & 0.8000          \\ \hline
\multicolumn{1}{|l|}{JNLP}                        & \multicolumn{1}{c|}{0.7526}            & \multicolumn{1}{c|}{0.6517}             & 0.8300          \\ \hline
\multicolumn{1}{|l|}{NOWJ}                        & \multicolumn{1}{c|}{0.7345}            & \multicolumn{1}{c|}{0.6892}             & 0.7750          \\ \hline
\multicolumn{1}{|l|}{HUKB}                        & \multicolumn{1}{c|}{0.6793}            & \multicolumn{1}{c|}{0.6342}             & 0.7150          \\ \hline
\multicolumn{1}{|l|}{LLNTU}                       & \multicolumn{1}{c|}{0.6600}            & \multicolumn{1}{c|}{0.7400}             & 0.6500          \\ \hline
\multicolumn{1}{|l|}{UA}                          & \multicolumn{1}{c|}{0.5698}            & \multicolumn{1}{c|}{0.6267}             & 0.5700          \\ \hline
\multicolumn{4}{|l|}{\textbf{Our system}}                                                                                                              \\ \hline
\multicolumn{1}{|l|}{gpt-4-1106-preview (4)}          & \multicolumn{1}{c|}{0.7647}            & \multicolumn{1}{c|}{\textbf{0.7564}}             & 0.7871          \\ \hline
\multicolumn{1}{|l|}{gpt-4-1106-preview-ensemble (5)} & \multicolumn{1}{c|}{\textbf{0.8085}}            & \multicolumn{1}{c|}{0.7277}             & \textbf{0.8712}          \\ \hline
\end{tabular}
\end{table}

Table \ref{tab:compare-res} presents the best runs of participating teams in the COLIEE 2023. The retrieval system (5) achieves an F2 score of 0.8085, which outperforms the best team in the competition by 4\%. Both our proposed retrieval systems using the GPT-4 model surpass participants in precision and recall scores. Indeed, these results emphasize the advantages of LLMs and prompting techniques in the retrieval task.

\subsection{Error Analysis}
\textbf{Overview}

\begin{figure}[ht]
    \includegraphics[width=0.75\linewidth]{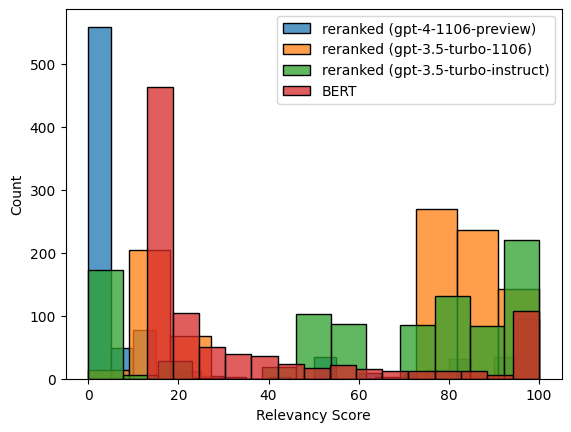}
    \caption{Relevancy score distribution on test set before and after re-ranking with different LLMs.}
    \label{fig:test-set-dist}
\end{figure}

\begin{table}
    \caption{Count and changes in precision, recall, and F2 by query.}
    \begin{tabular}{|c|c|c|c|} \hline 
         &  \textbf{Precision}& \textbf{Recall}&\textbf{F2}\\ \hline 
         Increased&  77 (avg. +0.6508)&  0&69 (avg. +0.4298)\\ \hline 
    Unchanged& 13& 81&13\\ \hline 
         Decreased&  11 (avg. -0.1253)&  20 (avg. -0.6749)&19 (avg. -0.2960)\\ \hline
    \end{tabular}
    \label{tab:score-changes}
\end{table}

Relevancy scores from BERT-based reranking and Prompting-based reranking's distribution of test set relevancy score are illustrated in \ref{fig:test-set-dist}. From the figure, it can be observed that gpt-3.5-turbo-1106 and gpt-3.5-turbo-instruct often give higher scores, eliminating fewer candidates. Meanwhile, gpt-4-1106-preview shares a similar distribution to BERT re-ranking, with most candidates eliminated with a score of 0.

Detailed changes in F2, precision, and recall per legal query are provided in Table \ref{tab:score-changes}. It is noticeable that Prompting Re-ranking often leads to drastic changes in both precision and recall, while it can help significantly increase precision in most cases, it comes with sacrificing an average of 0.67 recall in one-fifth of the test set. This highlights the limitations and trade-offs of this approach.

\textbf{Complicated samples (low precision and/or low recall)}

From outputs in the test set, this analysis aims to look for areas where LLMs struggle the most while re-ranking and discussing potential future research directions. The statistics in this section are taken from re-ranking outputs of only gpt-4-1106-preview without ensembling.

\textit{Many noisy candidates:} In cases where precision and recall are both 0, it is often due to the only matching articles getting eliminated, resulting in 0 precision and 0 recall at the same time. It seems that the model struggles to improve in cases where many noisy candidates are present. 14/101 samples were not able to recall any matching article, all of them have low precision and a high number of candidates from the previous stage, with up to 41 candidates for re-ranking averaging only 9\% precision. Around 3\% of samples already have 0 recall after the BERT re-ranking phase which LLMs re-ranking can not improve. To reduce noisy candidates from the BERT re-ranking phase, more improvements to the first two retrieval phases (BM25 and BERT re-ranking) are needed. This can be through techniques such as query extension, improved training regimes for the cross-encoder model, or other pre- and post-processing techniques. 

\textit{Complex legal situations with actors:} Half of the low recall queries (recall <= 0.5) or queries that saw recall decline after LLMs re-ranking are complex legal situations involving 2 or more actors. This highlights the challenges these models still face with complex reasoning and more techniques should be investigated to tackle this limitation.

\textbf{LLMs vs BERT}
\begin{figure}[ht]
    \begin{subfigure}[b]{0.49\textwidth}
        \centering
        \includegraphics[width=1\linewidth]{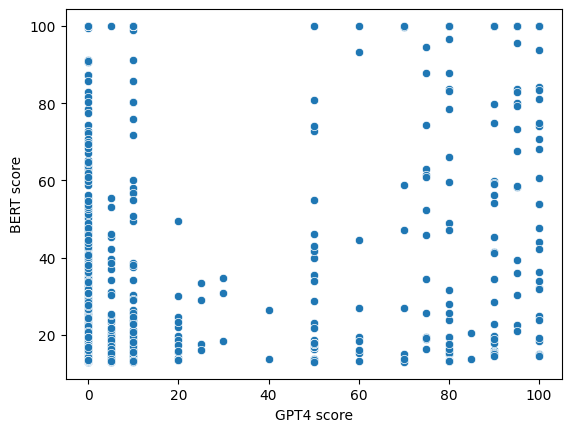}
        \caption{Scatterplot of GPT-4 scores against relevance scores derived from BERT-based re--ranking model.}
        \label{fig:scatter-gpt4}
    \end{subfigure}
    \begin{subfigure}[b]{0.49\textwidth}
        \centering
        \includegraphics[width=1\linewidth]{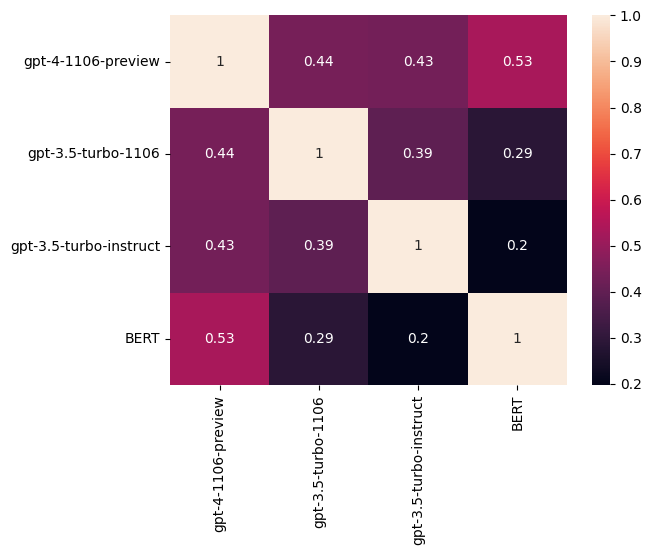}
        \caption{Correlation between models' scores.}
        \label{fig:correlation}
    \end{subfigure}
    \caption{Correlation Analysis}
\end{figure}

To further understand the differences between LLMs' outputs and BERT's, GPT4's scores are plotted against BERT's scores in \ref{fig:scatter-gpt4} and the correlation coefficient is calculated between models in \ref{fig:correlation}. Despite sharing a similar distribution as shown in \ref{fig:test-set-dist}, outputs from gpt-4-1106-preview and BERT show no significant correlation with Pearson correlation of only 0.532 between the two score sets.

This shows that GPT and BERT re-ranking models have notable differences in outputs and are good at different scenarios. This partly explains why when gpt-4-1106-preview and BERT are ensembled together, despite large F2 discrepancies shown in \ref{tab:experiment-result}, still lead to a sizable increase in F2—these 2 models when ensembled correct each other's biases hence leading to a more robust output.

\section{Conclusion}

The experiment in constructing a three-phase query system has demonstrated that, if implemented and utilized wisely, a large language model combined with prompting techniques can be fully leveraged to enhance the precision and recall in the search process. Simultaneously, analyzing query system errors has continued to uncover several unresolved weaknesses, including the complexity of queries and semantic interconnections among legal texts. These challenges will be the focal points of future research to develop a comprehensive solution for retrieval tasks in the legal data domain.

\section*{Acknowledgement}
Hai-Long Nguyen was funded by the Master, PhD Scholarship Programme of Vingroup Innovation Foundation (VINIF), code VINIF.2023.ThS.075. \\

\noindent This work was supported by JSPS KAKENHI Grant Numbers, JP22H00543 and JST, AIP Trilateral AI Research,Grant Number JPMJCR20G4.

\bibliographystyle{splncs04}
\bibliography{mybibliography}





\end{document}